\documentclass{article}

\usepackage[preprint]{neurips_2021}

\usepackage[utf8]{inputenc}
\usepackage[T1]{fontenc}
\usepackage{hyperref}
\usepackage{url}
\usepackage{algorithm} 
\usepackage[algo2e]{algorithm2e} 
\usepackage{array}
\usepackage{booktabs}
\usepackage{amsfonts}
\usepackage{amsmath}
\usepackage{amsthm}
\usepackage{enumerate}
\usepackage{graphicx}
\usepackage{multicol}
\usepackage{nicefrac}
\usepackage{microtype}
\usepackage{xcolor}

\usepackage{amsthm,amsmath,amsfonts,bm,xspace}
\usepackage{bbm}
\usepackage{color}
\usepackage{enumitem}

\def\eqref#1{(\ref{#1})}

\def\1{\bm{1}}

\def\vx{{\bm{x}}}

\DeclareMathAlphabet{\mathsfit}{\encodingdefault}{\sfdefault}{m}{sl}
\SetMathAlphabet{\mathsfit}{bold}{\encodingdefault}{\sfdefault}{bx}{n}

\newcommand{\E}{\mathbb{E}}

\newcommand{\KL}[2]{D_{\mathrm{KL}}\big(#1 \lVert #2\big)}

\newcolumntype{C}[1]{>{\centering\arraybackslash}m{#1}}

\title{Learning to Efficiently Sample from \\ Diffusion Probabilistic Models}

\author{
  Daniel Watson\thanks{Work done as part of the Google AI Residency.}\,,\, Jonathan Ho,\, Mohammad Norouzi,\, William Chan \\[.2cm]
  Google Research, Brain Team \\
  \texttt{\{watsondaniel,jonathanho,mnorouzi,williamchan\}@google.com} \\
}

\begin{document}

\maketitle

\begin{abstract}
  Denoising Diffusion Probabilistic Models (DDPMs) have emerged as a powerful family of generative models
  that can yield high-fidelity samples and competitive log-likelihoods across a range of domains, including image and speech synthesis.
  Key advantages of DDPMs include ease of training, in contrast to generative adversarial networks, and speed of generation, in contrast to autoregressive models. 
  However, DDPMs typically require hundreds-to-thousands of steps to generate a high fidelity sample, making them prohibitively expensive for high dimensional problems. 
  Fortunately, DDPMs allow trading generation speed for sample quality through adjusting the number of refinement steps as a post process.
  Prior work has been successful in improving generation speed through handcrafting the time schedule by trial and error.
  We instead view the selection of the inference time schedules as an optimization problem, and introduce an exact dynamic programming algorithm that finds the optimal discrete time schedules for any pre-trained DDPM.
  Our method exploits the fact that ELBO can be decomposed into separate KL terms, and given any computation budget, discovers the time schedule that maximizes the training ELBO exactly.
  Our method is efficient, has no hyper-parameters of its own, and can be applied to any pre-trained DDPM with no retraining.
  We discover inference time schedules requiring as few as 32 refinement steps, while sacrificing less than 0.1 bits per dimension compared to the default 4,000 steps used on ImageNet 64x64~\citep{ho2020denoising,nichol2021improved}.

\end{abstract}

\vspace{-.2cm}
\section{Introduction}
\vspace{-.1cm}

Denoising Diffusion Probabilistic Models (DDPMs) \citep{sohl2015deep,ho2020denoising} have emerged as
a powerful class of generative models,
which model the data distribution through an iterative denoising process.
DDPMs have been applied successfully to a variety of applications, including unconditional image generation \citep{song2019generative,ho2020denoising,song-iclr-2021,nichol2021improved}, shape generation \citep{cai-eccv-2020}, text-to-speech \citep{chen-iclr-2021, kong-arxiv-2020} and single image super-resolution \citep{saharia2021image, li2021srdiff}.

DDPMs are easy to train, featuring a simple denoising objective \citep{ho2020denoising} with noise schedules that successfully transfer across different models and datasets. This contrasts to Generative Adversarial Networks (GANs) \citep{goodfellow2014generative}, which require an inner-outer loop optimization procedure that often entails instability and requires careful hyperparameter tuning. DDPMs also admit a simple non-autoregressive inference process; this contrasts to autoregressive models with often prohibitive computational costs on high dimensional data. The DDPM inference process starts with samples from the corresponding prior noise distribution (e.g., standard Gaussian), and iteratively denoises the samples under the fixed noise schedule. However, DDPMs often need hundreds-to-thousands of denoising steps (each involving a feedforward pass of a large neural network) to achieve strong results. While this process is still much faster than autoregressive models, this is still often computationally prohibitive, especially when modeling high dimensional data.

There has been much recent work focused on improving the sampling speed of DDPMs. WaveGrad \citep{chen-iclr-2021} introduced a manually crafted schedule requiring only 6 refinement steps; however, this schedule seems to be only applicable to the vocoding task where there is a very strong conditioning signal. Denoising Diffusion Implicit Models (DDIMs) \citep{song2020denoising} accelerate sampling from pre-trained DDPMs by relying on a family of non-Markovian processes. They accelerate the generative process through taking multiple steps in the diffusion process. However, DDIMs sacrifice the ability to compute log-likelihoods.
\cite{nichol2021improved} also explored the use of ancestral sampling with a subsequence of the original denoising steps, trying both a uniform stride and other hand-crafted strides.
\cite{san2021noise} improve few-step sampling further by training a separate model after training a DDPM to estimate the level of noise, and modifying inference to dynamically adjust the noise schedule at every step to match the predicted noise level.

All these fast-sampling techniques rely on a key property of DDPMs -- there is a decoupling between the training and inference schedule. The training schedule need not be the same as the inference schedule, e.g., a diffusion model trained to use 1000 steps may actually use only 10 steps during inference. This decoupling characteristic is typically not found in other generative models. In past work, the choice of inference schedule was often considered a hyperpameter selection problem, and often selected via intuition or extensive hyperparmeter exploration \citep{chen-iclr-2021}.
In this work, we view the choice of inference schedule path as an independent optimization problem, wherein we attempt to learn the best schedule.
Our approach relies on a dynamic programming algorithm, where given a fixed budget of $K$ refinement steps and a pre-trained DDPM, we find the set of timesteps that maximizes the corresponding evidence lower bound (ELBO). As an optimization objective, the ELBO has a key \textit{decomposability} property: the total ELBO is the sum of individual KL terms, and for any two inference paths, if the timesteps $(s,t)$ contiguously occur in both, they share a common KL term, therefore admitting memoization.


Our main contributions are the following:
\vspace{-.1cm}
\begin{itemize}[topsep=0pt, partopsep=0pt, leftmargin=15pt, parsep=0pt, itemsep=1.75pt]    
    \item We introduce a dynamic programming algorithm that finds the optimal inference paths based on the ELBO for all possible computation budgets of $K$ refinement steps. The algorithm searches over $T>K$ timesteps, only requiring $\mathcal{O}(T)$  neural network forward passes. It only needs to be applied once to a pre-trained DDPM, does not require training or retraining a DDPM, and is applicable to both time-discrete and time-continuous DDPMs.
    \item We experiment with DDPM models from prior work. On both $L_\textrm{simple}$ CIFAR10 and $L_\textrm{hybrid}$ ImageNet 64x64, we discover schedules which require only 32 refinement steps, yet sacrifice only 0.1 bits per dimension compared to their original counterparts with 1,000 and 4,000 steps, respectively.
\end{itemize}

\vspace{-.1cm}
\section{Background on Denoising Diffusion Probabilistic Models}
\vspace{-.1cm}
\label{background}

Denoising Diffusion Probabilistic Models (DDPMs)~\citep{ho2020denoising,sohl2015deep} are defined in terms of a \textit{forward} Markovian diffusion process $q$ 
and a learned reverse process $p_\theta$. The forward diffusion process gradually adds Gaussian noise to a data point~$\vx_0$ through $T$ iterations,
\begin{eqnarray}
  q(\vx_{1:T} \mid \vx_0) &=& \prod\nolimits_{t=1}^{T} q(\vx_{t} \mid \vx_{t-1})~, \\
  q(\vx_{t} \mid \vx_{t-1}) &=& \mathcal{N}(\vx_{t} \mid \sqrt{\alpha_t}\, \vx_{t-1}, (1 - \alpha_t) \bm{I} )~, 
\end{eqnarray}
where the scalar parameters $\alpha_{1:T}$ determine the variance of the noise added at each diffusion step, subject to $0 < \alpha_t < 1$.
The learned reverse process aims to model $q(\vx_0)$ by inverting the forward process, gradually removing noise from signal
starting from pure Gaussian noise $\vx_T$,
\begin{eqnarray}
    p(\vx_T) &=& \mathcal{N}(\vx_T \mid \bm{0}, \bm{I}) \label{eq:mutheta}\\
    p_\theta(\vx_{0:T}) &=& p(\vx_T) \prod\nolimits_{t=1}^T p_\theta(\vx_{t-1} \mid \vx_t) \\
    p_\theta(\vx_{t-1} \mid \vx_{t}) &=& \mathcal{N}(\vx_{t-1} \mid \mu_{\theta}({\vx}_{t}, t), \sigma_t^2\bm{I})~. \label{eq:reverse_process}
\end{eqnarray}
The parameters of the reverse process can be optimized by maximizing the following variational lower bound on the training set,
\begin{equation}
    \E_{q} \log p(\vx_0) \ge \E_{q} \left[
    \log p_\theta(\vx_0 | \vx_1) - \!\sum_{t = 2}^T \KL{q(\vx_{t-1} | \vx_t, \vx_0)}{p_\theta(\vx_{t-1} | \vx_t)} - L_T(\vx_0)  \right] \label{eq:vlb}\!\!\!
\end{equation}
where $L_T(\vx_0) = \KL{q(\vx_T|\vx_0)\,}{\,p(\vx_T)}$. \cite{nichol2021improved} have demonstrated that training DDPMs by maximizing the ELBO yields competitive log-likelihood scores on both CIFAR-10 and ImageNet $64\!\times\!64$ achieving $2.94$ and $3.53$ bits per dimension respectively.

Two notable properties of Gaussian diffusion process that help formulate DDPMs tractably and efficiently include:
\begin{eqnarray}
    q(\vx_t \mid \vx_0) \!\!&\!\!=\!\!&\!\! \mathcal{N}(\vx_t \mid \sqrt{\gamma_t}\, \vx_0, (1-\gamma_t) \bm{I})~,\quad\quad \text{where}~\gamma_t = \prod\nolimits_{i=1}^t \alpha_i~,
    \label{eq:diffusion-marginal}\\
    \!\!\!\!\!\!\!\!q(\vx_{t-1} \mid \vx_0, \vx_t) \!\!&\!\!=\!\!&\!\!
    \mathcal{N}\!\left(\!\vx_{t-1} \,\Big|\, \frac{\sqrt{\gamma_{t-1}}\,(1-\alpha_t)\vx_0 + \sqrt{\alpha_t}\,(1-\gamma_{t-1})\vx_t}{1-\gamma_t}, \frac{(1-\gamma_{t-1})(1-\alpha_t)}{1-\gamma_t} \bm{I}\!\right).\!\!
    \label{eq:diffusion-posterior}
\end{eqnarray}
Given the marginal distribution of $\vx_t$ given $\vx_0$ in \eqref{eq:diffusion-marginal}, one can sample from the $q(\vx_t \mid \vx_0)$ independently for different $t$ and perform SGD on a randomly chosen KL term in \eqref{eq:vlb}.
Furthermore, given that the posterior distribution of $\vx_{t-1}$ given $\vx_{t}$ and $\vx_{0}$ is Gaussian, one can compute each KL term in \eqref{eq:vlb} between two Gaussians in closed form
and avoid high variance Monte Carlo estimation.

\vspace{-.1cm}
\section{Linking DDPMs to Continuous Time Affine Diffusion Processes}
\label{linking}
\vspace{-.1cm}

Before describing our approach to efficiently sampling from DDPMs, 
it is helpful to link DDPMs to continuous time {\em affine} diffusion processes, as it shows the compatibility of our approach to both time-discrete and time-continuous DDPMs. 
Let $\vx_0 \sim q(\vx_0)$ denote a data point drawn from the empirical distribution of interest and let $q(\vx_t|\vx_0)$ denote a stochastic process 
for $t \in [0,1]$ defined through an affine diffusion process through the following stochastic differential equation (SDE):
\begin{equation}
dX_t = f_{\textrm{sde}}(t) X_t dt + g_{\textrm{sde}}(t) dB_t~,
\end{equation}
where $f_{\textrm{sde}},g_{\textrm{sde}}: [0,1] \to [0,1]$ are integrable functions satisfying $f_{\textrm{sde}}(0) = 1$ and $g_{\textrm{sde}}(0) = 0$.

Following \cite{sarkka2019applied} (section 6.1), we can compute the exact marginals $q(\vx_t|\vx_s)$ for any $0\leq s < t \leq 1$. This differs from \cite{ho2020denoising}, where their marginals are those of the \textit{discretized} diffusion via Euler-Maruyama, where it is not possible to compute marginals outside the discretization since they are formulated as cumulative products. We get:
\begin{equation}
\label{eqn:sdebook}
q(\vx_t \mid \vx_s)
= \mathcal{N}\left(
  \vx_t \,\Big|\, \psi(t,s)\vx_s, \Big(\int_s^t \psi(t,u)^2 g(u)^2 du \Big) \bm{I}~
\right)
\end{equation}
where $\psi(t,s) = \exp\left(\int_s^t f(u) du\right)$. Since these integrals are difficult to work with, we instead propose to \textit{define} the marginals directly as
\begin{equation}
q(\vx_t \mid \vx_0) = \mathcal{N}(\vx_t \mid f(t)\vx_0, g(t)^2 \bm{I})~
\end{equation}
where $f,g: [0,1] \rightarrow [0,1]$ are differentiable, monotonic functions satisfying $f(0) = 1, f(1) = 0, g(0) = 0, g(1) = 1$.
Then, by implicit differentiation it follows that the corresponding diffusion is
\begin{equation}
\label{eqn:sde}
\begin{split}
    dX_t = \frac{f'(t)}{f(t)} X_t dt + \sqrt{2g(t)\left(g'(t) - \frac{f'(t)g(t)}{f(t)}\right)} dB_t~.
\end{split}
\end{equation}


To complete our formulation, let $f_{ts} = \frac{f(t)}{f(s)}$ and $g_{ts} = \sqrt{g(t)^2 - f_{ts}^2g(s)^2}$.
Then, it follows that for any $0 < s < t \leq 1$ we have that
\begin{eqnarray}
    q(\vx_t\mid\vx_s) &=& \mathcal{N}\left(\vx_t \mid f_{ts} \vx_s, g_{ts}^2 \bm{I} \right)~,
    \label{eqn:cont-marginal} \\
    q(\vx_s\mid\vx_t,\vx_0) &=& \mathcal{N}\left(\vx_s \,\Big|\, \frac{1}{g_{t0}^2}(f_{s0} g_{ts}^2 \vx_0 + f_{ts}g_{s0}^2 \vx_t), \frac{g_{s0}^2g_{ts}^2}{g_{t0}^2} \bm{I} \right)~,
    \label{eqn:cont-posterior}
\end{eqnarray}
Note that \eqref{eqn:cont-marginal} and \eqref{eqn:cont-posterior} can be thought of as generalizations of \eqref{eq:diffusion-marginal} and \eqref{eq:diffusion-posterior} to continuous time diffusion, i.e., this formulation not only includes that of \cite{ho2020denoising} as a special case, but also allows training DDPMs by sampling $t \sim \mathrm{Uniform}(0,1)$ like \cite{song-iclr-2021}, and is compatible with any choice of SDE (as opposed to \cite{song-iclr-2021} where one is limited to marginals and posteriors where the integrals in Equation \ref{eqn:sdebook} can be solved analytically). More importantly, we can also perform inference with \textit{any} ancestral sampling path (i.e., the timesteps can attain continuous values) by formulating the reverse process in terms of the posterior distribution as
\begin{equation}
p_\theta(\vx_{s}\mid \vx_{t})
= q\big(\vx_{s}\mid \vx_t, \hat{\vx}_0 = \tfrac{1}{f_{t 0}}(\vx_t - g_{t0} \epsilon_\theta(\vx_t,t))\big),
\end{equation}
justifying the compatibility of our main approach with time-continuous DDPMs.
We note that this reverse process is also mathematically equivalent to a reverse process based on a time-discrete DDPM derived from a subsequence of the original timesteps as done by \cite{song2020denoising,nichol2021improved}.

For the case of $s=0$ in the reverse process, we follow the parametrization of \cite{ho2020denoising} to obtain discretized log likelihoods and compare our log likelihoods fairly with prior work.


\section{Learning to Efficiently Sample from DDPMs}


We now introduce our dynamic programming (DP) approach. In general, after training a DDPM, there is a decoupling between training and inference schedules. One can use a different inference schedule compared to training. Additionally, we can optimize a loss or reward function with respect to the timesteps themselves (after the DDPM is trained). In this paper, we use the ELBO as our objective, however we note that it is possible to directly optimize the timesteps with other metrics. 

\subsection{Optimizing the ELBO}

In our work, we choose to optimize ELBO as our objective.
We rely on one key property of ELBO, its \textit{decomposability}.
We first make a few observations. The DDPM models the transition probability $p_\theta(x_s \mid x_t)$, or the cost to move from $x_t\rightarrow x_s$. Given a pretrained DDPM, one can construct any valid ELBO path through it as long as two properties hold:
\vspace{-.2cm}
\begin{enumerate}
    \item The path starts at $t=0$ and ends at $t=1$.
    \item The path is contiguously connected without breaks.
\end{enumerate}

We can construct an ELBO path that entails $K \in \mathbb{N}$ refinement steps. I.e., for any $K$, and any given path of inference timesteps $0 = t'_0 < t'_1 < ... < t'_{K-1} < t'_K = 1$, one can derive a corresponding ELBO
\begin{equation}
    -L_{\textrm{ELBO}}
    = \E_q \KL{q(\vx_1|\vx_0)}{p_\theta(\vx_1)} + \sum_{i=1}^K L(t'_i,t'_{i-1})
\end{equation}
where
\begin{equation}
\label{eqn:kl_term}
  L(t,s) = \begin{cases} 
    -\E_q \log p_\theta(\vx_t|\vx_0) & s = 0 \\
    \E_q \KL{q(\vx_s|\vx_t,\vx_0)}{p_\theta(\vx_s|\vx_t)} & s > 0
  \end{cases}
\end{equation}

In other words, the ELBO is a sum of individual ELBO terms that are functions of contiguous timesteps $(t'_i,t'_{i-1})$. Now the question remains, given a fixed budget $K$ steps, what is the optimal ELBO path?

First, we observe that any two paths that share a $(t, s)$ transition will share a common $L(t, s)$ term.
We exploit this property in our dynamic programming algorithm.
When given a grid of timesteps $0 = t_0 < t_1 < ... < t_{T-1} < t_T = 1$ with $T \geq K$, it is possible to efficiently find the exact optimum (i.e., finding $\{t'_1,...,t'_{K-1}\} \subset \{t_1,...,t_{T-1}\}$ with the best ELBO) by memoizing all the individual $L(t,s)$ ELBO terms for $s,t \in \{t_0,...,t_T\}$ with $s<t$. We can then solve the canonical least-cost-path problem on a directed graph where $s \to t$ are nodes and the edge connecting them has cost $L(t,s)$.

For time-continuous DDPMs, the choice of grid (i.e., the $t_1,...,t_{T-1}$) can be arbitrary. For models trained with discrete timesteps, the grid must be a subset of (or the full) original steps used during training, unless the model was regularized during training with methods such as the sampling procedure proposed by \cite{chen-iclr-2021}.

\begin{table*}[!tb]
  \begin{minipage}[t]{0.49\textwidth}
    \small
    \begin{algorithm}[H]
      \DontPrintSemicolon
      \textbf{input}: $L, T$ \quad\textcolor{gray}{\# L = KL cost table (Equation 17)} \; \\
      $D = \mathrm{np.full}((T+1,T+1), -1)$\; \\
      $C = \mathrm{np.full}((T+1,T+1), \mathrm{np.inf})$\; \\
      $C[0,0] = 0$\; \\
      \For{k $\mathrm{in}$ $\mathrm{range}(1, T + 1)$} {
        $bpds = C[k,\, $None$] + L[:, :]$\; \\
        $C[k] = \mathrm{np.amin}(bpds, \mathrm{axis=}-1$)\; \\
        $D[k] = \mathrm{np.argmin}(bpds, \mathrm{axis=}-1$)\;
      }
      \Return{D}\;
      \caption{Vectorized DP (all budgets)}
      \label{algo:dp_search}
    \end{algorithm}
  \end{minipage}
  \hfill
  \begin{minipage}[t]{0.49\textwidth}
    \small
    \begin{algorithm}[H]
      \DontPrintSemicolon
      \textbf{input}: $D, K$ \\
      $optpath = [\,]$\; \\
      $t = K$\; \\
      \For{k $\mathrm{in}$ $\mathrm{reversed(range(}(K))$} {
        $optpath.\mathrm{append}(t)$\; \\
        $t = D[k, t]$\;
      }
      \Return{optpath}\;
      \caption{Fetch shortest path of $K$ steps}
      \label{algo:dp_fetch}
    \end{algorithm}
  \end{minipage}
\end{table*}

\subsection{Dynamic Programming Algorithm}

We now outline our methodology to solve the least-cost-path problem. Our solution is similar to Dijkstra's algorithm, but it differs to the classical least-cost-path problem where the latter is typically used, as our problem has additional constraints: we restrict our search to paths of exactly $K + 1$ nodes, and the start and end nodes are fixed.


Let $C$ and $D$ be $(K+1) \times (T+1)$ matrices. $C[k,t]$ will be the total cost of the least-cost-path of length $k$ from $t$ to 0. $D$ will be filled with the timesteps corresponding to such paths; i.e., $D[k,t]$ will be the timestep $s$ immediately previous to $t$ for the optimal $k$-step path (assuming $t$ is also part of such path).

We initialize $C[0,0] = 0$ and all the other $C[0,\cdot]$ to $\infty$ (the $D[0,\cdot]$ are irrelevant, but for ease of index notation we keep them in this section). Then, for each $k$ from 1 to $K$, we iteratively set, for each $t$,
\begin{align*}
    C[k,t] &= \min_s \left(C[k-1,s] + L(t,s)\right) \\
    D[k,t] &= \arg\min_s \left(C[k-1,s] + L(t,s)\right)
\end{align*}
where $L(t,s)$ is the cost to transition from $t$ to $s$ (see Equation \ref{eqn:kl_term}). For all $s \geq t$, we set $L(t,s) = \infty$ (e.g., we only move backwards in the diffusion process). This procedure captures the shortest path cost in $C$ and the shortest path itself in $D$.


We further observe that running the DP algorithm for each $k$ from 1 to $T$ (instead of $K$), we can extract the optimal paths for \textit{all} possible budgets $K$. Algorithm 1 illustrates a vectorized version of the procedure we have outlined in this section, while Algorithm 2 shows how to explicitly extract the optimal paths from $D$. 

\subsection{Efficient Memoization}

A priori, our dynamic programming approach appears to be inefficient because it requires computing $\mathcal{O}(T^2)$ terms (recall, as we rely on all the $L(t,s)$ terms which depend on a neural network forward pass). We however observe that a single forward pass of the DDPM can be used to compute \textit{all} the $L(t,\cdot)$ terms. This holds true even in the case where the pre-trained DDPM \textit{learns} the variances. For example, in \cite{nichol2021improved} instead of fixing them to $\Tilde{g}_{ts}$ as we outlined in the previous section, the forward pass itself still only depends on $t$ and not $s$, and the variance of $p_\theta(x_s|x_t)$ is obtained by interpolating the forward pass's output logits $\bm{v}$ with $\exp(\bm{v} \log g_{ts}^2 + (1 - \bm{v}) \log \Tilde{g}_{ts}^2)$. Thus, computing the table of all the $L(t,s)$ ELBO terms only requires $\mathcal{O}(T)$ forward passes.


\section{Experiments}
\label{experiments}

We apply our method on a wide variety of pre-trained DDPMs from prior work. This emphasizes the fact that our method is applicable to any pre-trained DDPM model. In particular, we rely the CIFAR10 model checkpoints released by \cite{nichol2021improved} on both their $L_{\textrm{hybrid}}$ and $L_{\textrm{vlb}}$ objectives. We also showcase results on CIFAR10 \citep{krizhevsky2009learning} with the exact configuration used by \cite{ho2020denoising}, which we denote as $L_{\textrm{simple}}$, as well as $L_{\textrm{hybrid}}$ on ImageNet 64x64 \citep{deng2009imagenet} following \cite{nichol2021improved}, training these last two models ourselves for 800K and 3M steps, respectively, but otherwise using the exact same configurations as the authors.

\begin{table}[!tb]
\centering
\caption{Negative log likelihoods (bits/dim) in the few-step regime across various DDPMs trained on CIFAR10, as well as state-of-the-art unconditional generative models in the same dataset. The last column corresponds to 1,000 steps for $L_{\textrm{simple}}$ and 4,000 steps for all other models.}
\begin{tabular}{ p{6.5cm}||m{.525cm}|m{.525cm}|m{.525cm}|m{.525cm}|m{.525cm}|m{.525cm}|m{.575cm} }
 \hline
 Model $\setminus$ \# refinement steps & 8 & 16 & 32 & 64 & 128 & 256 & All \\
 \hline
 \hline
 DistAug Transformer \citep{jun20distaug} & -- & -- & -- & -- & -- & -- & \textbf{2.53} \\
 \hline
 DDPM++ (deep, sub-VP) \citep{song-iclr-2021} & -- & -- & -- & -- & -- & -- & \textbf{2.99} \\
 \hline
 $L_{\textrm{simple}}$ & & & & & & &  \\
 \quad Even stride & 6.95 & 6.15 & 5.46 & 4.91 & 4.47 & 4.14 & 3.73 \\
 \quad Quadratic stride & 5.39 & 4.86 & 4.52 & 3.84 & 3.74 & 3.73 &  \\
 \quad DP stride & 4.59 & 3.99 & 3.79 & 3.74 & 3.73 & 3.72 &  \\
 \hline
 $L_{\textrm{vlb}}$ & & & & & & &  \\
 \quad Even stride & 6.20 & 5.48 & 4.89 & 4.42 & 4.03 & 3.73 & \textbf{2.94} \\
 \quad Quadratic stride & 4.89 & 4.09 & 3.58 & 3.23 & 3.09 & 3.05 &  \\
 \quad DP stride & 4.20 & 3.41 & 3.17 & 3.08 & 3.05 & 3.04 &  \\
 \hline
 $L_{\textrm{hybrid}}$ & & & & & & &  \\
 \quad Even stride & 6.14 & 5.39 & 4.77 & 4.29 & 3.92 & 3.66 & 3.17 \\
 \quad Quadratic stride & 4.91 & 4.15 & 3.71 & 3.42 & 3.30 & 3.26 &  \\
 \quad DP stride & 4.33 & 3.62 & 3.39 & 3.30 & 3.27 & 3.26 &  \\
 \hline
\end{tabular}
\end{table}

\begin{table}[!tb]
\centering
\caption{Negative log likelihoods (bits/dim) in the few-step regime for a DDPM model trained with $L_{\textrm{hybrid}}$ on ImageNet 64x64 \citep{nichol2021improved}, as well as state-of-the-art unconditional generative models in the same dataset. We underline that, with just 32 steps, our DP stride achieves a score of $\le 0.1$ bits/dim higher than the same model with the original 4,000 step budget ($^*$the authors report 3.57 bits/dim, but we trained the model for 3M rather than 1.5M steps).}
\begin{tabular}{ p{6.5cm}||m{.525cm}|m{.525cm}|m{.525cm}|m{.525cm}|m{.525cm}|m{.525cm}|m{.575cm} }
 \hline
 Model $\setminus$ \# refinement steps & 8 & 16 & 32 & 64 & 128 & 256 & 4000 \\
 \hline
 \hline
 Routing Transformer \citep{roy2021efficient} & -- & -- & -- & -- & -- & -- & \textbf{3.43} \\
 \hline
 $L_{\textrm{vlb}}$ \citep{nichol2021improved} & -- & -- & -- & -- & -- & -- & \textbf{3.53} \\
 \hline
 $L_{\textrm{hybrid}}$ & & & & & & & \\
 \quad Even stride & 6.07 & 5.38 & 4.82 & 4.39 & 4.08 & 3.87 & $\textbf{3.55}^*$ \\
 \quad Quadratic stride & 4.83 & 4.14 & 3.82 & 3.65 & 3.58 & 3.56 &  \\
 \quad DP stride & 4.29 & 3.80 & \underline{3.65} & 3.59 & 3.56 & 3.56 &  \\
 \hline
\end{tabular}
\end{table}

In our experiments, we always search over a grid that includes all the timesteps used to train the model, i.e., $\{t/T : t \in \{1,...,T-1\}\}$. For our CIFAR10 results, we computed the memoization tables with Monte Carlo estimates over the full training dataset, while on ImageNet 64x64 we limited the number of datapoints in the Monte Carlo estimates to 16,384 images on the training dataset.

\begin{figure}[htp]
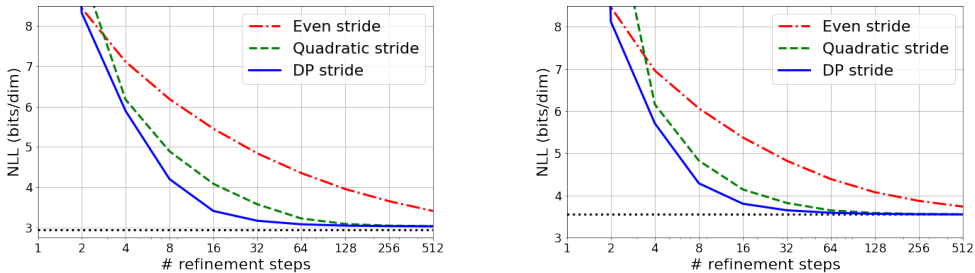

    \centering
    \begin{minipage}{0.49\textwidth}
        \centering
        \includegraphics[width=0.99\textwidth]{"plots/vlb_bpd_corrected"}
    \end{minipage}\hfill
    \begin{minipage}{0.49\textwidth}
        \centering
        \includegraphics[width=0.99\textwidth]{"plots/imagenet_bpd_corrected"}
    \end{minipage}
    \caption{Negative log likelihoods (bits/dim) for $L_{\textrm{vlb}}$ CIFAR10 (left) and $L_{\textrm{hybrid}}$ ImageNet 64x64 (right) for strides discovered via dynamic programming v.s. even and quadratic strides.}
\end{figure}

For each pre-trained model, we compare the negative log likelihoods (estimated using the full heldout dataset) of the strides discovered by our dynamic programming algorithm against even and quadratic strides, following \cite{song2020denoising}. We find that our dynamic programming algorithm discovers strides resulting in much better log likelihoods than the hand-crafted strides used in prior work, particularly in the few-step regime. We provide a visualization of the log likelihood curves as a function of computation budget in Figure 1 for $L_{\textrm{simple}}$ CIFAR10 and $L_{\textrm{hybrid}}$ ImageNet 64x64 \citep{deng2009imagenet}, a full list of the scores in the few-step regime in Table 1, and a visualization of the discovered steps themselves in Figure 2.

\subsection{Comparison with FID}

We further evaluate our discovered strides by reporting FID scores \citep{heusel2017gans} on 50,000 model samples against the same number of samples from the training dataset, as is standard in the literature. We find that, although our strides are yield much better log likelihoods, such optimization does not necessarily translate to also improving the FID scores. Results are included in Figure 3. This weakened correlation between log-likehoods and FID is consistent with observations in prior work \citep{ho2020denoising,nichol2021improved}.

\begin{figure}[!tp]
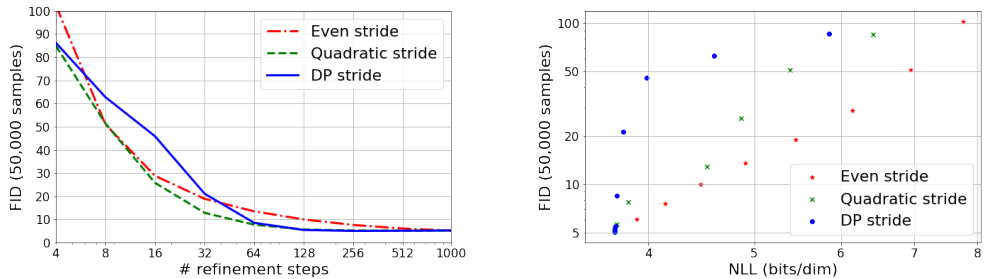

    \centering
    \begin{minipage}{0.49\textwidth}
        \centering
        \includegraphics[width=0.99\textwidth]{"plots/simple_fid"}
    \end{minipage}\hfill
    \begin{minipage}{0.49\textwidth}
        \centering
        \includegraphics[width=0.99\textwidth]{"plots/simple_bpd_fid_comp"}
    \end{minipage}
    \caption{FID scores for $L_{\textrm{simple}}$ CIFAR10, as a function of computation budget (left) and negative log likelihood (right).}
\end{figure}

\subsection{Monte Carlo Ablation}
\label{subsection:montecarlo}

To investigate the feasibility of our approach using minimal computation, we experimented with setting the number of Monte Carlo datapoints used to compute the dynamic programming table of negative log likelihood terms to 128 samples (i.e., easily fit into a single batch of GPU memory). We find that, for CIFAR10, the difference in log likelihoods is negligible, while on ImageNet 64x64 there is a visible yet slight improvement in negative log likelihood when filling the table with more samples. We hypothesize that this is due to the higher diversity of ImageNet. Nevertheless, we highlight that our procedure can be applied very quickly (i.e., with just $T$ forward passes of a neural network when using a single batch, as opposed to a running average over batches), even for large models, to significantly improve log their likelihoods in the few-step regime.

\begin{figure}[htp]
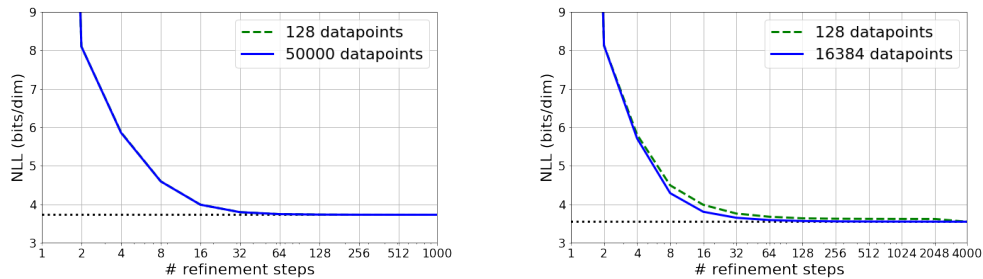

    \centering
    \begin{minipage}{0.49\textwidth}
        \centering
        \includegraphics[width=0.99\textwidth]{"plots/simple_mc_abl"}
    \end{minipage}\hfill
    \begin{minipage}{0.49\textwidth}
        \centering
        \includegraphics[width=0.99\textwidth]{"plots/imagenet_mc_abl"}
    \end{minipage}
    \caption{Negative log likelihoods (bits/dim) for $L_{\textrm{simple}}$ CIFAR10 and $L_{\textrm{hybrid}}$ ImageNet 64x64 for strides discovered via dynamic programming with log-likelihood term tables estimated with a varying number of datapoints.}
\end{figure}

\begin{figure}[htp]
\small
    \centering
    \begin{minipage}{0.49\textwidth}
        \centering
        
        
        32 steps
        \includegraphics[width=0.99\textwidth]{"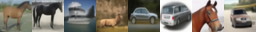"}
        \includegraphics[width=0.99\textwidth]{"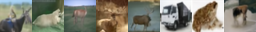"}
        \includegraphics[width=0.99\textwidth]{"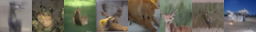"}
        
        128 steps
        \includegraphics[width=0.99\textwidth]{"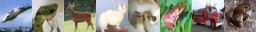"}
        \includegraphics[width=0.99\textwidth]{"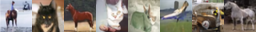"}
        \includegraphics[width=0.99\textwidth]{"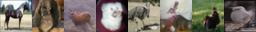"}
        
        1,000 steps
        \includegraphics[width=0.99\textwidth]{"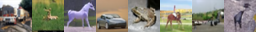"}
    \end{minipage}\hfill
    \begin{minipage}{0.49\textwidth}
        \centering

        
        64 steps
        \includegraphics[width=0.99\textwidth]{"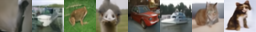"}
        \includegraphics[width=0.99\textwidth]{"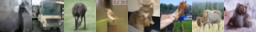"}
        \includegraphics[width=0.99\textwidth]{"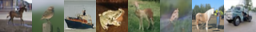"}
        
        256 steps
        \includegraphics[width=0.99\textwidth]{"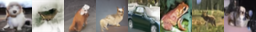"}
        \includegraphics[width=0.99\textwidth]{"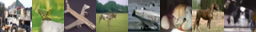"}
        \includegraphics[width=0.99\textwidth]{"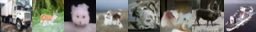"}
        
        Real samples
        \includegraphics[width=0.99\textwidth]{"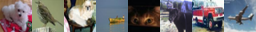"}
    \end{minipage}
    \caption{Non-cherrypicked $L_{\textrm{simple}}$ CIFAR10 samples for even (top), quadratic (middle), and DP  strides (bottom), for various computation budgets.
    Samples are based on the same 8 random seeds.}
\end{figure}

\begin{figure}[t]
\small
    \centering
    \begin{minipage}{0.5\textwidth}
        \centering
        
        
        32 steps
        \includegraphics[width=0.99\textwidth]{"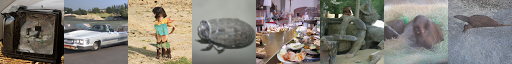"}
        \includegraphics[width=0.99\textwidth]{"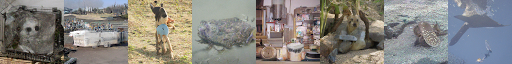"}
        \includegraphics[width=0.99\textwidth]{"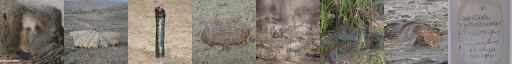"}
        
        128 steps
        \includegraphics[width=0.99\textwidth]{"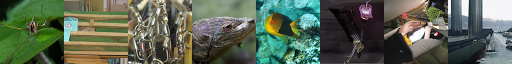"}
        \includegraphics[width=0.99\textwidth]{"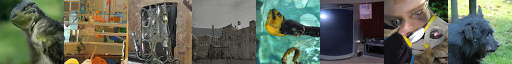"}
        \includegraphics[width=0.99\textwidth]{"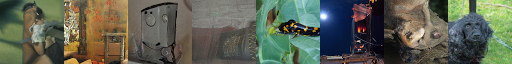"}
        
        4,000 steps
        \includegraphics[width=0.99\textwidth]{"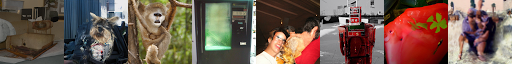"}
    \end{minipage}\hfill
    \begin{minipage}{0.5\textwidth}
        \centering

        
        64 steps
        \includegraphics[width=0.99\textwidth]{"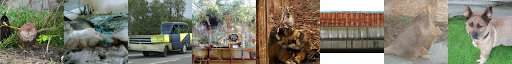"}
        \includegraphics[width=0.99\textwidth]{"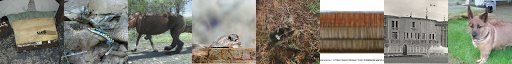"}
        \includegraphics[width=0.99\textwidth]{"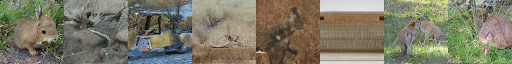"}
        
        256 steps
        \includegraphics[width=0.99\textwidth]{"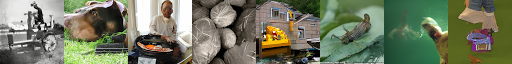"}
        \includegraphics[width=0.99\textwidth]{"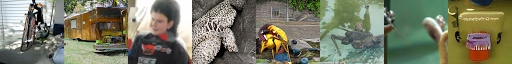"}
        \includegraphics[width=0.99\textwidth]{"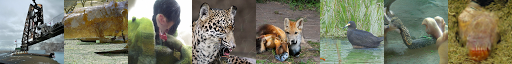"}
        
        Real samples
        \includegraphics[width=0.99\textwidth]{"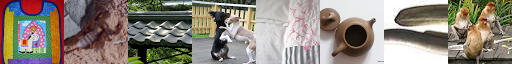"}
    \end{minipage}
    \caption{Non-cherrypicked $L_{\textrm{hybrid}}$ ImageNet 64x64 samples for even (top), quadratic (middle), and DP strides (bottom), for various computation budgets. Samples are based on the same 8 random seeds.}
\end{figure}



\begin{figure}[htp]
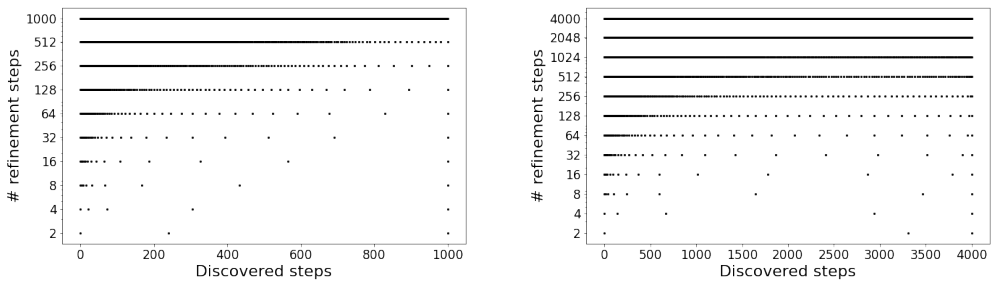

    \centering
    \begin{minipage}{0.495\textwidth}
        \centering
        \includegraphics[width=1.0\textwidth]{"plots/simple_paths"}
    \end{minipage}\hfill
    \begin{minipage}{0.495\textwidth}
        \centering
        \includegraphics[width=1.0\textwidth]{"plots/imagenet_paths"}
    \end{minipage}
    \caption{Timesteps discovered via dynamic programming for $L_{\textrm{simple}}$ CIFAR10 (left) and $L_\textrm{hybrid}$ ImageNet 64x46 (right) for various computation budgets. Each step (forward pass) is between two contiguous points. Our DP algorithm prefers allocates steps towards the end of the diffusion, agreeing with intuition from prior work where steps closer to $\vx_0$ are important as they capture finer image details, but curiously, it may also allocate steps closer to $\vx_1$, possibly to better break modes early on in the diffusion process.}
\end{figure}

\section{Related Work}

DDPMs \citep{ho2020denoising} have recently shown results that are competitive with GANs \citep{goodfellow2014generative}, and they can be traced back to the work of \cite{sohl2015deep} as a restricted family of deep latent variable models. \cite{dhariwal2021diffusion} have more recently shown that DDPMs can outperform GANs in FID scores \citep{heusel2017gans}.
\cite{song2019generative} have also linked DDPMs to denoising score matching \citep{vincent2008extracting,vincent2010stacked}, which is crucial to the continuous-time formulation \citep{song-iclr-2021}. This connection to score matching has been explored further by \cite{song2021train}, where other score-matching techniques (e.g., sliced score matching, \cite{song2020sliced}) have been shown to be valid DDPM objectives and DDPMs are linked to energy-based models.
More recent work on the few-step regime of DDPMs \citep{song2020denoising,chen-iclr-2021,nichol2021improved,san2021noise,kong2021fast,jolicoeurmartineau2021gotta} has also guided our research efforts. DDPMs are also very closely related to variational autoencoders \citep{Kingma2013}, where more recent work has shown that, with many stochastic layers, they can also attain competitive negative log likelihoods in unconditional image generation \citep{child2020very}. Also very closely related to DDPMs, there has also been work on non-autoregressive modeling of text sequences that can be regarded as discrete-space DDPMs with a forward process that masks or remove tokens \citep{lee2018deterministic,gu-neurips-2019,stern-icml-2019,chan-icml-2020,saharia-arxiv-2020}. The UNet architecture \citep{ronneberger2015u} has been key to the recent success of DDPMs, and as shown by \cite{ho2020denoising,nichol2021improved}, augmenting UNet with self-attention \citep{shaw2018self} in scales where attention is computationally feasible has helped bring DDPMs closer to the current state-of-the-art autoregressive generative models \citep{child2019generating,jun20distaug,roy2021efficient}.

\section{Conclusion and Discussion}
\label{discussion}

By regarding the selection of the inference schedule as an optimization problem, we present a novel and efficient dynamic programming algorithm to discover the optimal inference schedule for a pre-trained DDPM. Our DP algorithm finds an optimal inference schedule based on the ELBO given a fixed computation budget. Our method need only be applied once to discover the schedule, and does not require training or re-training the DPPM. In the few-step regime, we discover schedules on $L_\textrm{simple}$ CIFAR10 and $L_\textrm{hybrid}$ ImageNet 64x64 that require only 32 steps, yet sacrifice $\le 0.1$ bits per dimension compared to state-of-the-art DDPMs using hundreds-to-thousands of refinement steps. Our approach only needs forward passes of the DDPM neural network to fill the dynamic programming table of $L(t,s)$ terms, and we show that we can fill the dynamic programming table with just $\mathcal{O}(T)$ forward passes. Moreover, we show that we can estimate the table using only 128 Monte Carlo samples, finding this to be sufficient even for datasets such as ImageNet with high diversity. Our method achieves strong likelihoods with very few refinement steps, outperforming prior work utilizing hand-crafted strides \citep{ho2020denoising,nichol2021improved}. 

Despite very strong log-likelihood results, especially in the few step regime, we observe limitations to our method. There is a disconnect between log-likehoods and FID scores, where improvements in log-likelihoods do not necessarily translate to improvements in FID scores.
This is consistent with prior work, showing that the correlation between log likelihood and FID can be mismatched \citep{ho2020denoising,nichol2021improved}.
We hope our work will encourage future research exploiting our general framework of optimization post-training in DDPMs, potentially utilizing gradient-based optimization over not only the ELBO, but also other, non-decomposable metrics. We particularly note that other sampling steps such as MCMC corrector steps or alternative predictor steps (e.g., following the reverse SDE) \citep{song-iclr-2021} can also be incorporated into computation budget, and general learning frameworks like reinforcement learning are well-suited to explore this space as well as non-differentiable learning signals.




\clearpage

\bibliographystyle{plainnat}
\bibliography{neurips_2021}

\begin{thebibliography}{36}
\providecommand{\natexlab}[1]{#1}
\providecommand{\url}[1]{\texttt{#1}}
\expandafter\ifx\csname urlstyle\endcsname\relax
  \providecommand{\doi}[1]{doi: #1}\else
  \providecommand{\doi}{doi: \begingroup \urlstyle{rm}\Url}\fi

\bibitem[Cai et~al.(2020)Cai, Yang, Averbuch-Elor, Hao, Belongie, Snavely, and
  Hariharan]{cai-eccv-2020}
Ruojin Cai, Guandao Yang, Hadar Averbuch-Elor, Zekun Hao, Serge Belongie, Noah
  Snavely, and Bharath Hariharan.
\newblock {Learning Gradient Fields for Shape Generation}.
\newblock In \emph{{ECCV}}, 2020.

\bibitem[Chan et~al.(2020)Chan, Saharia, Hinton, Norouzi, and
  Jaitly]{chan-icml-2020}
William Chan, Chitwan Saharia, Geoffrey Hinton, Mohammad Norouzi, and Navdeep
  Jaitly.
\newblock {Imputer: Sequence Modelling via Imputation and Dynamic Programming}.
\newblock In \emph{{ICML}}, 2020.

\bibitem[Chen et~al.(2021)Chen, Zhang, Zen, Weiss, Norouzi, and
  Chan]{chen-iclr-2021}
Nanxin Chen, Yu~Zhang, Heiga Zen, Ron~J. Weiss, Mohammad Norouzi, and William
  Chan.
\newblock {WaveGrad: Estimating Gradients for Waveform Generation}.
\newblock In \emph{{ICLR}}, 2021.

\bibitem[Child(2020)]{child2020very}
Rewon Child.
\newblock Very deep vaes generalize autoregressive models and can outperform
  them on images.
\newblock \emph{arXiv preprint arXiv:2011.10650}, 2020.

\bibitem[Child et~al.(2019)Child, Gray, Radford, and
  Sutskever]{child2019generating}
Rewon Child, Scott Gray, Alec Radford, and Ilya Sutskever.
\newblock Generating long sequences with sparse transformers.
\newblock \emph{arXiv preprint arXiv:1904.10509}, 2019.

\bibitem[Deng et~al.(2009)Deng, Dong, Socher, Li, Li, and
  Fei-Fei]{deng2009imagenet}
Jia Deng, Wei Dong, Richard Socher, Li-Jia Li, Kai Li, and Li~Fei-Fei.
\newblock Imagenet: A large-scale hierarchical image database.
\newblock In \emph{2009 IEEE conference on computer vision and pattern
  recognition}, pages 248--255. Ieee, 2009.

\bibitem[Dhariwal and Nichol(2021)]{dhariwal2021diffusion}
Prafulla Dhariwal and Alex Nichol.
\newblock Diffusion models beat gans on image synthesis.
\newblock \emph{arXiv preprint arXiv:2105.05233}, 2021.

\bibitem[Goodfellow et~al.(2014)Goodfellow, Pouget-Abadie, Mirza, Xu,
  Warde-Farley, Ozair, Courville, and Bengio]{goodfellow2014generative}
Ian~J Goodfellow, Jean Pouget-Abadie, Mehdi Mirza, Bing Xu, David Warde-Farley,
  Sherjil Ozair, Aaron Courville, and Yoshua Bengio.
\newblock Generative adversarial networks.
\newblock \emph{arXiv preprint arXiv:1406.2661}, 2014.

\bibitem[Gu et~al.(2019)Gu, Wang, and Zhao]{gu-neurips-2019}
Jiatao Gu, Changhan Wang, and Jake Zhao.
\newblock {Levenshtein Transformer}.
\newblock In \emph{{NeurIPS}}, 2019.

\bibitem[Heusel et~al.(2017)Heusel, Ramsauer, Unterthiner, Nessler, and
  Hochreiter]{heusel2017gans}
Martin Heusel, Hubert Ramsauer, Thomas Unterthiner, Bernhard Nessler, and Sepp
  Hochreiter.
\newblock Gans trained by a two time-scale update rule converge to a local nash
  equilibrium.
\newblock \emph{arXiv preprint arXiv:1706.08500}, 2017.

\bibitem[Ho et~al.(2020)Ho, Jain, and Abbeel]{ho2020denoising}
Jonathan Ho, Ajay Jain, and Pieter Abbeel.
\newblock {Denoising Diffusion Probabilistic Models}.
\newblock \emph{{NeurIPS}}, 2020.

\bibitem[Jolicoeur-Martineau et~al.(2021)Jolicoeur-Martineau, Li,
  Piché-Taillefer, Kachman, and Mitliagkas]{jolicoeurmartineau2021gotta}
Alexia Jolicoeur-Martineau, Ke~Li, Rémi Piché-Taillefer, Tal Kachman, and
  Ioannis Mitliagkas.
\newblock Gotta go fast when generating data with score-based models, 2021.

\bibitem[Jun et~al.(2020)Jun, Child, Chen, Schulman, Ramesh, Radford, and
  Sutskever]{jun20distaug}
Heewoo Jun, Rewon Child, Mark Chen, John Schulman, Aditya Ramesh, Alec Radford,
  and Ilya Sutskever.
\newblock {Distribution Augmentation for Generative Modeling}.
\newblock In \emph{{ICML}}, 2020.

\bibitem[Kingma and Welling(2013)]{Kingma2013}
Diederik~P Kingma and Max Welling.
\newblock {Auto-Encoding Variational Bayes}.
\newblock In \emph{ICLR}, 2013.

\bibitem[Kong and Ping(2021)]{kong2021fast}
Zhifeng Kong and Wei Ping.
\newblock On fast sampling of diffusion probabilistic models, 2021.

\bibitem[Kong et~al.(2020)Kong, Ping, Huang, Zhao, and
  Catanzaro]{kong-arxiv-2020}
Zhifeng Kong, Wei Ping, Jiaji Huang, Kexin Zhao, and Bryan Catanzaro.
\newblock {DiffWave: A Versatile Diffusion Model for Audio Synthesis}.
\newblock \emph{arXiv preprint arXiv:2009.09761}, 2020.

\bibitem[Krizhevsky et~al.(2009)Krizhevsky, Hinton,
  et~al.]{krizhevsky2009learning}
Alex Krizhevsky, Geoffrey Hinton, et~al.
\newblock Learning multiple layers of features from tiny images.
\newblock \emph{Technical Report}, 2009.

\bibitem[Lee et~al.(2018)Lee, Mansimov, and Cho]{lee2018deterministic}
Jason Lee, Elman Mansimov, and Kyunghyun Cho.
\newblock Deterministic non-autoregressive neural sequence modeling by
  iterative refinement.
\newblock \emph{arXiv preprint arXiv:1802.06901}, 2018.

\bibitem[Li et~al.(2021)Li, Yang, Chang, Feng, Xu, Li, and Chen]{li2021srdiff}
Haoying Li, Yifan Yang, Meng Chang, Huajun Feng, Zhihai Xu, Qi~Li, and Yueting
  Chen.
\newblock {SRDiff: Single Image Super-Resolution with Diffusion Probabilistic
  Models}.
\newblock \emph{arXiv:2104.14951}, 2021.

\bibitem[Nichol and Dhariwal(2021)]{nichol2021improved}
Alex Nichol and Prafulla Dhariwal.
\newblock Improved denoising diffusion probabilistic models.
\newblock \emph{arXiv preprint arXiv:2102.09672}, 2021.

\bibitem[Ronneberger et~al.(2015)Ronneberger, Fischer, and
  Brox]{ronneberger2015u}
Olaf Ronneberger, Philipp Fischer, and Thomas Brox.
\newblock U-net: Convolutional networks for biomedical image segmentation.
\newblock In \emph{International Conference on Medical image computing and
  computer-assisted intervention}, pages 234--241. Springer, 2015.

\bibitem[Roy et~al.(2021)Roy, Saffar, Vaswani, and Grangier]{roy2021efficient}
Aurko Roy, Mohammad Saffar, Ashish Vaswani, and David Grangier.
\newblock Efficient content-based sparse attention with routing transformers.
\newblock \emph{Transactions of the Association for Computational Linguistics},
  9:\penalty0 53--68, 2021.

\bibitem[Saharia et~al.(2020)Saharia, Chan, Saxena, and
  Norouzi]{saharia-arxiv-2020}
Chitwan Saharia, William Chan, Saurabh Saxena, and Mohammad Norouzi.
\newblock {Non-Autoregressive Machine Translation with Latent Alignments}.
\newblock \emph{{EMNLP}}, 2020.

\bibitem[Saharia et~al.(2021)Saharia, Ho, Chan, Salimans, Fleet, and
  Norouzi]{saharia2021image}
Chitwan Saharia, Jonathan Ho, William Chan, Tim Salimans, David~J Fleet, and
  Mohammad Norouzi.
\newblock Image super-resolution via iterative refinement.
\newblock \emph{arXiv preprint arXiv:2104.07636}, 2021.

\bibitem[San-Roman et~al.(2021)San-Roman, Nachmani, and Wolf]{san2021noise}
Robin San-Roman, Eliya Nachmani, and Lior Wolf.
\newblock Noise estimation for generative diffusion models.
\newblock \emph{arXiv preprint arXiv:2104.02600}, 2021.

\bibitem[S{\"a}rkk{\"a} and Solin(2019)]{sarkka2019applied}
Simo S{\"a}rkk{\"a} and Arno Solin.
\newblock \emph{Applied stochastic differential equations}, volume~10.
\newblock Cambridge University Press, 2019.

\bibitem[Shaw et~al.(2018)Shaw, Uszkoreit, and Vaswani]{shaw2018self}
Peter Shaw, Jakob Uszkoreit, and Ashish Vaswani.
\newblock Self-attention with relative position representations.
\newblock \emph{arXiv preprint arXiv:1803.02155}, 2018.

\bibitem[Sohl-Dickstein et~al.(2015)Sohl-Dickstein, Weiss, Maheswaranathan, and
  Ganguli]{sohl2015deep}
Jascha Sohl-Dickstein, Eric Weiss, Niru Maheswaranathan, and Surya Ganguli.
\newblock Deep unsupervised learning using nonequilibrium thermodynamics.
\newblock In \emph{International Conference on Machine Learning}, pages
  2256--2265. PMLR, 2015.

\bibitem[Song et~al.(2020{\natexlab{a}})Song, Meng, and
  Ermon]{song2020denoising}
Jiaming Song, Chenlin Meng, and Stefano Ermon.
\newblock Denoising diffusion implicit models.
\newblock \emph{arXiv preprint arXiv:2010.02502}, 2020{\natexlab{a}}.

\bibitem[Song and Ermon(2019)]{song2019generative}
Yang Song and Stefano Ermon.
\newblock {Generative Modeling by Estimating Gradients of the Data
  Distribution}.
\newblock \emph{{NeurIPS}}, 2019.

\bibitem[Song and Kingma(2021)]{song2021train}
Yang Song and Diederik~P Kingma.
\newblock How to train your energy-based models.
\newblock \emph{arXiv preprint arXiv:2101.03288}, 2021.

\bibitem[Song et~al.(2020{\natexlab{b}})Song, Garg, Shi, and
  Ermon]{song2020sliced}
Yang Song, Sahaj Garg, Jiaxin Shi, and Stefano Ermon.
\newblock Sliced score matching: A scalable approach to density and score
  estimation.
\newblock In \emph{Uncertainty in Artificial Intelligence}, pages 574--584.
  PMLR, 2020{\natexlab{b}}.

\bibitem[Song et~al.(2021)Song, Sohl-Dickstein, Kingma, Kumar, Ermon, and
  Poole]{song-iclr-2021}
Yang Song, Jascha Sohl-Dickstein, Diederik~P. Kingma, Abhishek Kumar, Stefano
  Ermon, and Ben Poole.
\newblock {Score-Based Generative Modeling through Stochastic Differential
  Equations}.
\newblock In \emph{{ICLR}}, 2021.

\bibitem[Stern et~al.(2019)Stern, Chan, Kiros, and Uszkoreit]{stern-icml-2019}
Mitchell Stern, William Chan, Jamie Kiros, and Jakob Uszkoreit.
\newblock {Insertion Transformer: Flexible Sequence Generation via Insertion
  Operations}.
\newblock In \emph{{ICML}}, 2019.

\bibitem[Vincent et~al.(2008)Vincent, Larochelle, Bengio, and
  Manzagol]{vincent2008extracting}
Pascal Vincent, Hugo Larochelle, Yoshua Bengio, and Pierre-Antoine Manzagol.
\newblock Extracting and composing robust features with denoising autoencoders.
\newblock In \emph{Proceedings of the 25th international conference on Machine
  learning}, pages 1096--1103, 2008.

\bibitem[Vincent et~al.(2010)Vincent, Larochelle, Lajoie, Bengio, Manzagol, and
  Bottou]{vincent2010stacked}
Pascal Vincent, Hugo Larochelle, Isabelle Lajoie, Yoshua Bengio, Pierre-Antoine
  Manzagol, and L{\'e}on Bottou.
\newblock Stacked denoising autoencoders: Learning useful representations in a
  deep network with a local denoising criterion.
\newblock \emph{Journal of machine learning research}, 11\penalty0 (12), 2010.

\end{thebibliography}

\clearpage

\appendix

\section{Appendix}

\subsection{Proof for Equation \ref{eqn:sde}}

From Equation \ref{eqn:sdebook}, we get by implicit differentiation that
\begin{align*}
& f(t) = \psi(t,0) = \exp\left(\int_0^t f_{\textrm{sde}}(u) du\right) \\
\Rightarrow & f'(t) = \exp\left(\int_0^t f_{\textrm{sde}}(u) du\right) \frac{d}{dt} \int_0^t f_{\textrm{sde}}(u) du
             = f(t) f_{\textrm{sde}}(t) \\
\Rightarrow & f_{\textrm{sde}}(t) = \frac{f'(t)}{f(t)}
\end{align*}

Similarly as above and also using the fact that $\psi(t,s) = \frac{\psi(t,0)}{\psi(s,0)}$,

\begin{align*}
& g(t)^2 = \int_0^t \psi(t,u)^2 g_{\textrm{sde}}(u)^2 du
         = \int_0^t \frac{f(t)^2}{f(u)^2} g_{\textrm{sde}}(u)^2 du
         = f(t)^2\int_0^t \frac{g_{\textrm{sde}}(u)^2}{f(u)^2} du \\
\Rightarrow & 2g(t)g'(t) = 2f(t)f'(t) \frac{g(t)^2}{f(t)^2} + f(t)^2 \frac{d}{dt} \int_0^t \frac{g_{\textrm{sde}}(u)^2}{f(u)^2} du
                         = 2f_{\textrm{sde}}(t)g(t)^2 + g_{\textrm{sde}}(t)^2 \\
\Rightarrow & g_{\textrm{sde}}(t) = \sqrt{2(g(t)g'(t) - f_{\textrm{sde}}(t)g(t)^2)}. \qed
\end{align*}

\subsection{Proof for Equations \ref{eqn:cont-marginal} and \ref{eqn:cont-posterior}}

From Equation \ref{eqn:sdebook} and $\psi(t,s) = \frac{\psi(t,0)}{\psi(s,0)}$ it is immediate that $f_{ts}$ is the mean of $q(x_t|x_s)$. To show that $g_{ts}^2$ is the variance of $q(x_t|x_s)$, Equation \ref{eqn:sdebook} implies that
\begin{align*}
\mathrm{Var}[x_t|x_s] &= \int_s^t \psi(t,u)^2 g_{\textrm{sde}}(u)^2 du \\
                      &= \int_0^t \psi(t,u)^2 g_{\textrm{sde}}(u)^2 du - \int_0^s \psi(t,u)^2 g_{\textrm{sde}}(u)^2 du \\
                      &= g(t)^2 - \psi(t,0)^2 \int_0^s \frac{\psi(s,u)^2}{\psi(s,u)^2\psi(u,0)^2} g_{\textrm{sde}}(u)^2 du \\
                      &= g(t)^2 - \psi(t,0)^2 \int_0^s \frac{\psi(s,u)^2}{\psi(s,0)^2} g_{\textrm{sde}}(u)^2 du \\
                      &= g(t)^2 - \psi(t,s)^2 g(s)^2 \\
                      &= g(t)^2 - f_{ts} g(s)^2.
\end{align*}

The mean of $q(x_s|x_t,x_0)$ is given by the Gaussian conjugate prior formula (where all the distributions are conditioned on $x_0$). Let $\mu = f_{ts}x_s$, so we have a prior over $\mu$ given by
\[
x_s|x_0 \sim \mathcal{N}(f_{s0}x_0,g_{s0}^2 I_d)
\Rightarrow
\mu|x_0 \sim \mathcal{N}(f_{s0}f_{ts} x_0,f_{ts}^2 g_{s0}^2 I_d) \sim \mathcal{N}(f_{t0} x_0,f_{ts}^2 g_{s0}^2 I_d),
\]
and a likelihood with mean $\mu$
\[
x_t|x_s,x_0 \sim x_t|x_s \sim \mathcal{N}(f_{ts}x_s,g_{ts}^2 I_d)
\Rightarrow
x_t|\mu,x_0 \sim x_t|\mu \sim \mathcal{N}(\mu,g_{ts}^2 I_d).
\]
Then it follows by the formula that $\mu|x_t,x_0$ has variance
\begin{align*}
& \mathrm{Var}[\mu|x_t,x_0]
= \left(\frac{1}{f_{ts}^2 g_{s0}^2} + \frac{1}{g_{ts}^2} \right)^{-1}
= \left(\frac{g_{ts}^2 + f_{ts}^2 g_{s0}^2}{f_{ts}^2 g_{s0}^2 g_{ts}^2} \right)^{-1}
= \frac{f_{ts}^2 g_{s0}^2 g_{ts}^2}{g_{ts}^2 + f_{ts}^2 g_{s0}^2} \\
\Rightarrow
& \mathrm{Var}[x_s|x_t,x_0]
= \frac{1}{f_{ts}^2} \mathrm{Var}[\mu|x_t,x_0]
= \frac{g_{s0}^2 g_{ts}^2}{g_{ts}^2 + f_{ts}^2 g_{s0}^2}
= \frac{g_{s0}^2 g_{ts}^2}{g_{t0}^2}
= \Tilde{g}_{ts}^2
\end{align*}

and mean
\begin{align*}
& \mathbb{E}[\mu|x_t,x_0]
= \left(\frac{1}{f_{ts}^2 g_{s0}^2} + \frac{1}{g_{ts}^2} \right)^{-1} \left(\frac{f_{t0} x_0}{f_{ts}^2 g_{s0}^2} + \frac{x_t}{g_{ts}^2} \right)
= \frac{f_{t0}g_{ts}^2 x_0 + f_{ts}^2 g_{s0}^2 x_t}{g_{ts}^2 + f_{ts}^2 g_{s0}^2}
= \frac{f_{t0}g_{ts}^2 x_0 + f_{ts}^2 g_{s0}^2 x_t}{g_{t0}^2}\\
\Rightarrow
& \mathbb{E}[x_s|x_t,x_0]
= \frac{1}{f_{ts}} \mathbb{E}[\mu|x_t,x_0]
= \frac{\cfrac{f_{t0}}{f_{ts}} g_{ts}^2 x_0 + f_{ts} g_{s0}^2 x_t}{g_{t0}^2}
= \frac{f_{s0}g_{ts}^2 x_0 + f_{ts} g_{s0}^2 x_t}{g_{t0}^2}
= \Tilde{f}_{ts}(x_t,x_0). \qed
\end{align*}

\end{document}